\documentclass{article}

\usepackage[final]{nips_2018}

\usepackage[utf8]{inputenc} 
\usepackage[T1]{fontenc}    
\usepackage{hyperref}       
\usepackage{url}            
\usepackage{booktabs}       
\usepackage{amsfonts}       
\usepackage{nicefrac}       
\usepackage{microtype}      
\usepackage{amssymb}  
\usepackage{amsthm}
\usepackage{graphicx}
\usepackage{amsmath}
\usepackage{xcolor}
\title{Pre-training Graph Neural Networks with Kernels}

\author{
  Nicolò Navarin \\
  Department of Mathematics\\
  University of Padova, Italy\\
  \texttt{nnavarin@math.unipd.it} \\
  \And
   Dinh V. Tran \\
   Department of Computer Science \\
   University of Freiburg, Germany \\
  \texttt{dinh@informatik.uni-freiburg.de} \\
   \And
   Alessandro Sperduti \\
  Department of Mathematics\\
  University of Padova, Italy\\
  \texttt{sperduti@math.unipd.it} \\
}

\begin{document}

\maketitle

\begin{abstract}
  Many machine learning techniques have been proposed in the last few years to process data represented in graph-structured form. Graphs can be used to model several scenarios, from molecules and materials to RNA secondary structures. 
  Several kernel functions have been defined on graphs that coupled with kernelized learning algorithms, have shown state-of-the-art performances on many tasks.
  Recently, several definitions of Neural Networks for Graph (GNNs) have been proposed, but their accuracy is not yet satisfying.
  In this paper, we propose a task-independent pre-training methodology that allows a GNN to learn the representation induced by state-of-the-art graph kernels. Then, the supervised learning phase will fine-tune this representation for the task at hand.
  The proposed technique is agnostic on the adopted GNN architecture and kernel function, and shows consistent improvements in the predictive performance of GNNs in our preliminary experimental results.

\end{abstract}

\section{Introduction}
Chemists often rely on virtual screening for the design of new drugs and materials. Being the space of chemical compounds huge, Machine Learning (ML) models can help searching this space.
For instance, it is possible to design a ML task as the classification of toxic vs. non-toxic compounds, or predicting if a drug will be active or not against a certain disease, or if a compound will bind to a specific protein. In this way, chemists can focus on just the subset of molecules predicted as positive from the ML algorithm.

State-of-the-art machine learning techniques for classification and regression on graphs are at the moment kernel machines equipped with specifically designed kernels for graphs (e.g,
~\cite{shervashidze2009efficient,vishwanathan2010graph,MartinoNS12}). Although there are examples of kernels for structures that can be designed on the basis of a training set \citep{Fisher, sombased, GenerativeKerneles}, most of the more efficient and effective graph kernels are based on predefined structural features, i.e, features definition is not part of the learning process.  

There is a recent shift of trend from kernels to neural networks for graphs. Unlike kernels, the  features in neural networks are defined based on a learning process which is supervised by the graph's labels (targets).
Many approaches have addressed the problem of defining neural networks for graphs~\citep{surveyNN4G}.
However, training such networks is difficult, and collecting labels for big datasets is expensive, if even feasible.

A commonly accepted technique to improve the performances of machine learning models without the need for large labeled datasets is unsupervised pre-training. In many applications involving graphs, we have a big number of unlabeled examples available (e.g. we know the chemical structure of hundreds of thousands of chemical compounds, but we don't have labeled data for just a small subset of them for a specific learning task). Many techniques have been proposed to adopt the regularities in unlabeled data to drive the training phase and improve the generalization performance of ML models.
However, to date, we are not aware of pre-training techniques specifically designed for inputs represented as graphs.

Researchers are already trying to integrate the knowledge acquired from the design of kernels in neural networks \citep{Lei2017}.
In this paper, we propose a pre-training techniques for Graph Neural Networks that exploits the representations induced by graph kernels to drive the learning phase.
In a nutshell, our proposal is to define a siamese network for the pre-training stage that, fixed a graph kernel, and given a pair of graphs in input, computes an approximation of the kernel value.
When such a network converges, the learned representation is going to be close to the one induced by the graph kernel. Then, we fine-tune the representation via back-propagation on the labeled training set.
\section{Background}
In this section, we review all the basic components of our proposed approach.
\subsection{Definitions and notations}
\label{sec:def}
We denote matrices with bold uppercase letters, vectors with uppercase letters, and variables with lowercase letters. Given a matrix $\mathbf{M}$, $M_i$ denotes the $i$-th row of the matrix, and $m_{ij}$ is the element in $i$-th row and $j$-th column. Given the vector $V$, $v_i$ refers its $i$-th element. 

Let us consider $G=(V^G, E^G, \mathbf{X}^G)$ as a graph, where \mbox{$V^G=\{v_1, \ldots, v_n\}$} is the set of vertices (or nodes), \mbox{$E_G \subseteq V^G \times V^G$} is the set of edges, and $\mathbf{X}^G \in \mathbb{R}^{n\times d}$ is a node label matrix, where each row is the label (a vector of size $d$) associated to each vertex $v_i \in V^G$ , i.e. \mbox{$X^G_i = (x_{i,1}, \ldots, x_{i,d})$}. Note that, in this paper, we will not consider edge labels.
When the reference to the graph $G$ is clear from the context, for the sake of notation we discard the superscript referring to the specific graph. 
We define the adjacency matrix $A \in \mathbb{R}^{n \times n}$ as ${{a}}_{ij}=1 \iff (i,j) \in E$, 0 otherwise.
We also define the \emph{neighborhood} of a vertex $v$ as the set of vertices connected to $v$ by an edge, i.e. $\mathcal{N}(v)=\{u | (v,u) \in E\}$. Note that $N(v)$ is also the set of nodes at shortest path distance exactly one from $v$, i.e. $\mathcal{N}(v)=\{u | sp(v,u)=1\}$, where $sp$ is a function computing the shortest-path distance between two nodes in a graph.

In this paper, we consider the problem of graph classification. Given a dataset composed of $N$ pairs \mbox{$\{(x_i, y_i) | 1 \leq i \leq N\}$}, the task is then, given an unseen graph $x$, to predict its correct target $y$.
We will consider, for this learning task, graph neural networks and graph kernels, that are discussed in the next sections.
\subsection{Graph Kernels}
\label{sec:graphkernels}
A kernel on $\cal{X}$, the input space, is a symmetric positive semi-definite function $k: \cal{X}\times \cal{X} \rightarrow \mathbb{R}$ computing a score (\textit{similarity}) between pairs of instances. Kernel functions compute the dot product between two objects in a \emph{Reproducing Kernel Hilbert Space} (RKHS), i.e.: \mbox{$k(x,y)= \langle \phi(x), \phi(y) \rangle$} where \mbox{$\phi: \cal{X} \rightarrow \cal{H}$} is a function mapping instances from $\cal{X}$ to the RKHS (or feature space) $\cal{H}$. Different kernels define different feature spaces.

In this section, we describe several graph kernels proposed in literature, in terms of the graph substructures that they considers as features in thier respective \textit{RKHS}s.
As we will discuss later, every graph kernel can be adopted in the pre-training phase of our proposed method (see Section~\ref{sec:proposal}).

Random walk kernels (MGK) \citep{Gartner2003a} considers  as features all the possible random walks in a graph.

The Shortest Path (SP) Kernel associates a feature to each pair of node labels at a certain (unbounded) shortest-path distance \citep{Kriegel05shortestpath}.

\cite{Vert2009} describe a graph kernel based on extracting tree patterns from the graph. 
A much more efficient kernel based on tree patterns is the Weisfeiler-Lehman (WL) Subtree Kernel that counts the number of identical subtree patterns obtained by breadth-first visits where each node can appear multiple times \cite{Shervashidze2011}. The kernel depends on an hyper-parameter $h$, that is the a-priori selected number of WL iterations, that corresponds to the maximum depth of the considered patterns.
%
\cite{Costa2010} extended the WL Subtree Kernel by computing exact matches between pairs of subgraphs with controlled size and distance.

A family of graph kernels based on visits, i.e. the  Ordered Decomposition DAGs kernels,
has been proposed in \cite{DaSanMartino2012,DaSanMartino2016}. 
The framework has also been extended to consider graphs with continuous attributes \citep{DaSanMartino2017}. We do not consider continuous attributes in this work, however it is one of its natural extensions and we plan to consider such type of graphs in an extended version of this paper.

More recent frameworks for graph kernels addressing efficiency or efficacy have been proposed in \cite{Neumann2015,Orsini}.

\subsection{Neural Networks for Graphs}
\label{sec:graphconvolution}
The core machine learning models that we are going to adopt in this paper are neural networks for graphs. Our proposed pre-training method can work, in principle, with all the models presented in this section.

 The first 
 definition of neural network for graphs has been proposed in~\cite{SperdutiStarita}. More recent models have been 
proposed in \cite{Micheli2009,Scarselli2009}.
Both works are based on an idea that has been re-branded later as \emph{graph convolution}.

The idea is to define the neural architecture following the topology of the graph. Then a transformation is performed from the neurons corresponding to a vertex and its neighborhood to a hidden representation, that is associated to the same vertex (possibly in another layer of the network).
This transformation depends on some parameters, that are shared among all the nodes.
In the following, for the sake of simplicity we ignore the bias terms. 

In \cite{Scarselli2009}, a transition function on a graph node $v$ at time $0\leq t$ is defined as:
\begin{equation}
H^{t+1}_v =\sum_{u \in \mathcal{N}(v)} f_{\Theta}({H}^{t}_u,{X}_v,{X}_u),
\label{eq:scarsellirec}
\end{equation}
where $f_\Theta$ is a parametric function whose parameters $\Theta$ have to be learned (e.g. a neural network) and are shared among all the vertices.
Note that, if edge labels are available, they can be included in eq.~\eqref{eq:scarsellirec}. In fact, in the original formulation, $f_{\Theta}$ depends also on the label of the edge between $v$ and $u$.
This transition function is part of a recurrent system. It is defined as a contraction mapping, thus the system is guaranteed to converge to a fixed point, i.e. a representation, that does not depend on the particular initialization of the weight matrix $\mathbf{H}^0$.
The output is computed from the representation in the last layer, and from the original node labels as follows:
\begin{equation}
O^t_v=g_{\Theta'}(H^t_v,X_v),
\end{equation}
where $g_{\Theta'}$ is another neural network.
\cite{Li2015b} extends the work in \cite{Scarselli2009} by removing the constraint for the recurrent system to be a contraction mapping, and replacing the recurrent units with GRUs.
However, recently it has been shown in~\cite{Bresson2018} that stacked graph convolutions are superior to graph recurrent architectures in terms of both accuracy and computational cost.

In \cite{Micheli2009}, a model referred to as Neural Network for Graphs (NN4G) is proposed. In the first layer, a transformation over node labels is computed:
\begin{equation}
\hat{h}^1_v=f \left ( \sum_{j=1}^{d} \bar{w}_{1,j} x_{v,j} \right ),
\label{eq:micheliconv1}
\end{equation}
where $\bar{W}_1$ are the weights connecting the original labels $X$ to the current neuron, and $1 \leq v \leq n$ is the vertex index.
The graph convolution is then defined for the $i+1$-th layer (for $i>0$) as:
\begin{equation}
\hat{h}^{i+1}_v=f \left ( \sum_{j=1}^{d} \bar{w}_{i+1,j} x_{v,j}+
\sum_{k=1}^{i} \hat{w}_{i+1,k}\sum_{u \in \mathcal{N}(v)} \hat{h}^k_u
\right ),
\label{eq:micheliconv}
\end{equation}
where $\hat{W}_{i+1}$ are weights connecting the previous hidden layers to the current neuron (shared).
Note that in this formulation, skip connections are present, to the $(i+1)$-th layer, from layer $1$ to layer $i$. There is an interesting recent work about the parallel between skip-connections (residual networks in that case) and recurrent networks~\citep{Liao2016}. However, since in the formulation in eq.~\eqref{eq:micheliconv}, every layer is connected to all the subsequent layers, it is not possible to reconduct it to a (vanilla) recurrent model.
Let us consider the \mbox{$(i+1)$-th} graph convolutional layer, that comprehends $c_{i+1}$ graph convolutional filters. We can rewrite eq.~\eqref{eq:micheliconv} for the whole layer as:
\begin{equation}
\mathbf{H}^{i+1}= f(\mathbf{X} \mathbf{\bar{W}}^{i+1} + \sum_{k=1}^{i}\mathbf{A} \mathbf{H}^k \mathbf{\hat{W}}^{i+1,k}),
\label{eq:micheliconvmatrix}
\end{equation}
where ${i={0,\ldots,l-1}}$ (and $l$ is the number of layers), \mbox{$\mathbf{\bar{W}}^{i+1} \in \mathbb{R}^{d \times c_{i+1}}$}, \mbox{$\mathbf{\hat{W}}^{{i+1},k} \in \mathbb{R}^{c_{k} \times c_{i+1}}$}, $\mathbf{H}^k \in \mathbb{R}^{n \times c_k}$, $c_i$ is the size of the hidden representation at the $i$-th layer, and $f$ is applied element-wise.

The convolution in eq.~\eqref{eq:micheliconv} is part of a multi-layer architecture, where each layer's connectivity resembles the topology of the graph, and the training is layer-wise. Finally, for each graph, NN4G computes the average graph node representation for each hidden layer, and concatenates them. This is the graph representation computed by \textit{NN4G}, and it can be used for the final prediction of graph properties with a standard output layer.

In \cite{Duvenaud2015b}, a hierarchical approach has been proposed. This method is similar to NN4G and is inspired by circular fingerprints in chemical structures. While \cite{Micheli2009} adopts Cascade-Correlation for training, \cite{Duvenaud2015b} uses an end-to-end back-propagation.
 \cite{simonovsky2017dynamic} propose an improvement of \cite{Duvenaud2015b}, weighting the sum over the neighbors of a node by weights conditioned by the edge labels. 

Recently, \cite{Kipf2016a} derived a graph convolution that closely resembles eq.~\eqref{eq:micheliconv}. Let us, from now on, consider $\mathbf{H}^0=\mathbf{X}$.
Motivated by a first-order approximation of localized spectral filters on graphs, the proposed graph convolutional filter looks like:
\begin{equation}
\mathbf{H}^{i+1}= f(\mathbf{\tilde{D}}^{-\frac{1}{2}} \mathbf{\tilde{A}} \mathbf{\tilde{D}}^{-\frac{1}{2}} \mathbf{H}^i \mathbf{W}^i),
\label{eq:kipf}
\end{equation}
where $\mathbf{\tilde{A}}=\mathbf{A}+\mathbf{I}$, $\tilde{d}_{ii} = \sum_j \tilde{{a}}_{i,j}$, and $f$ is any activation function applied element-wise.

If we ignore the terms $\mathbf{\tilde{D}}^{-\frac{1}{2}}$ (that in practice act as normalization), it is easy to see that eq.~\eqref{eq:kipf} is very similar to  eq.~\eqref{eq:micheliconvmatrix}, the difference being that there are no skip connections in this case, i.e. the $(i+1)$-th layer is connected to the $i$-th layer only. Consequently, we just have to learn one weight matrix per layer.

\label{sec:dgcnn}
In \cite{Zhang2018}, a slightly more complex model compared to \cite{Kipf2016a} is proposed. This model shows the highest predictive performance with respect to the other methods presented in this section. The first layers of the network are again stacked graph convolutional layers, defined as follows:
\begin{equation}
\textbf{H}^{i+1}=f(\mathbf{\tilde{D}}^{-1}\mathbf{\tilde{A}} \mathbf{H}^{i} \mathbf{W}^{i}),
\label{eq:dgcnn}
\end{equation}
where $\mathbf{H}^0=\mathbf{X}$ and $\mathbf{\tilde{A}}=\mathbf{A}+\mathbf{I}$. Note that in the previous equation, we compute the representation of all the nodes in the graph at once.
Both equations~\eqref{eq:kipf} and~\eqref{eq:dgcnn}  can be seen as  first-order approximations of the polynomially parameterized spectral graph convolution.
In \cite{Zhang2018}, three graph convolutional layers are stacked.
The graph convolutions are followed by a concatenation layer that merges the representations computed by each graph convolutional layer. Then, differently from previous approaches, the paper introduces a \emph{sortpooling} layer, that selects a fixed number of node representations, and computes the output from them stacking 1D convolutional layers and dense layers.
This is the  network architecture that we considered in this paper, that will be referred to as DGCNN.
An alternative to the \emph{sortpooling} layer has been adopted in \cite{pmlr-v70-gilmer17a}.

Recently in \cite{pgcdgcnn2018}, the model in \cite{Zhang2018} has been extended. A parametric convolution operation have been proposed, that improves the expressiveness and the predictive performance of the graph convolution operation.
We plan to experiment with this other model in a future version of this work.

\subsection{Siamese Networks}
Siamese networks have been introduced in \cite{BROMLEY1993}, originally for the task of signature identification. 
In classical machine learning problems, e.g. in supervised learning, the goal is to predict a label for each example in a dataset.
However, not all machine learning problems are defined on a single example.
In particular, in some cases the labels are associated to pairs of examples.
More formally, in this setting, a training set is in the form $\{(x^{(i)}_1, x^{(i)}_2, y^{(i)}), i=1, \ldots, n\}$, where $x^{(i)}_j \in \mathcal{X}$ and $y^{(i)} \in \mathcal{Y}$.
The learning task is then to compute a function $\mathcal{X} \times \mathcal{X} \rightarrow \mathcal{Y}$.
Siamese networks apply the same function $f(x)$ to the two inputs, providing an encoding for each input (i.e. $f: \mathcal{X} \rightarrow \mathbb{R}^d$).
The output $y$ then depends on some other function $g(f(x_1),f(x_2))$. If both $f$ and $g$ are neural networks, it is possible to learn the two functions together via back-propagation.

\section{Related works: pre-training for deep Neural Networks}
The breakthrough that allowed to effectively train large-scale “deep” networks has been the introduction of an unsupervised pre-training phase \citep{Hinton2006}, in which the network is trained to build a generative model of the data, which can be subsequently refined using a supervised criterion (fine-tuning phase). Pre-training initializes the weights of the network in a region where optimization is somehow easier, thus helping the fine-tuning phase to reach better local optima. It might also perform some form of regularization, by introducing a bias towards good configurations of the parameter space \cite{Erhan2010}.
Pre-training for structured data has a more recent history when concerning sequences, while not much has been proposed up to now for more complex structures, such as trees and graphs.
Concerning sequences, earlier approaches to pre-training for sequences did not take into account temporal dependencies (e.g., \cite{Mohamed2012,BoulangerLewandowski2012ModelingTD}) only involving input-to-hidden connections, i.e. assuming  each item belonging to a sequence as independent of the others. 
This pre-training strategy is clearly unsatisfactory, because by definition the items belonging to the same sequence are dependent on each other, and this information should be exploited also during pre-training.
More effective strategies have been proposed more recently. Specifically,
in \cite{Pasa2014a},
a pre-training technique for recurrent neural networks based on linear autoencoder networks for sequences has been proposed. The idea is to use the weights obtained by the (approximated)  closed form solution  of a linear autoencoder, given a training set of sequences, as initial weights for the input-to-hidden connections of a recurrent neural network, which is then trained on the desired supervised task. 
A more general approach, i.e. independent from the specific architecture of the model, has been proposed in \cite{Pasa2015}. It is based on the training of a first-order HMM to generate an approximate distribution of the original dataset of sequences. The generative model is then used to generate a ``smoothed'' (reduced in size) dataset for pre-training, using the learning procedure associated with the chosen model. Starting from the weights obtained by the pre-training procedure, learning is then finalized by a fine-tuning phase where the original dataset is used.

Concerning neural networks for more complex structures, such as graphs that we consider in this paper, at the best of our knowledge there is no pre-training approach proposed in literature.

\section{Kernel-based pre-training}
\label{sec:proposal}
When dealing with machine learning on graphs, kernel methods are among the methods of choice, due to the theoretical guarantees they provide, and to the many efficient and effective kernel functions proposed in literature in the last years (see Section~\ref{sec:graphkernels}).
Such kernels are known to provide a good representation for the input graphs, showing state-of-the-art results when coupled with linear models (in the feature space) such as SVM.
%
We recall that kernels generally compute the representations only implicitly (since the feature space can be very high - or even infinite- dimensional), since kernel functions actually compute the dot product between the representations of two inputs.

On the other hand, given the relatively small size of graph datasets available online (some thousand examples, compared to the millions of examples usually required for training very deep neural networks),
over-parametrized methods such as graph neural networks are hard to train, providing often solutions that do not generalize well.
In this context, the pre-training phase may significantly impact the predictive performance of the Graph Neural Network.

One possibility for pre-training is to force the network to learn a fixed and good representation for graphs from the considered domain. After pre-training, the network will start from a set of parameters that already provide good performance - as opposed to random initialization. Such representation may be provided by graph kernels.
However, explicit representations in graph kernels' feature spaces usually are very high dimensional, thus training a network to reproduce it is a difficult task.

We propose a pre-training method to use graph kernels' representations to drive the learning process of a Graph Neural Network, without resorting to its explicit feature space.
In doing that, we aim to be as efficient and general as possible.
In a nutshell, our proposal is to define a siamese network, that, fixed a graph kernel, and given a pair of graphs in input, computes an approximation of the kernel value.

Let us consider a graph dataset $D=\{(x^{(i)},y^{(i)}, i=1,\ldots,N\}$, where $x^{(i)} \in \mathcal{G}$ (the space of graphs), and a set of unlabeled examples $U=\{x^{(i)}, i=1, \ldots, M\}$ with examples from the same domain $x^{(i)} \in \mathcal{G}$ (possibly $D \subseteq U$). Suppose that we would like to train a graph neural network for the task described by $D$.
We can exploit the information in $U$ with a pre-training approach.
Our method consists in defining a 
siamese network on all the possible pairs of examples in $U$. The output of the network is a single real value, and we define it as the values of a kernel function between the two inputs, i.e. $y=g(f(x_1), f(x_2))=k(x_1,x_2)$ for some choice of $k$.
The dataset that we use for the pre-training siamese network is then $\{(x^{(i)},x^{(j)},k(x^{(i)},x^{(j)}), i,j = 1,\ldots,M  \}$ for $x^{(i)} \in U$.

If we define $g$ (the output layer of the siamese network) as a function that computes the dot product between its inputs, i.e. $g(f(x_1), f(x_2)) = \langle f(x_1), f(x_2) \rangle$, then we are forcing the network $f(\cdot)$ to learn a compressed representation of the kernel's feature space.

\begin{figure}[t]
    \centering
    \includegraphics[width=\textwidth]{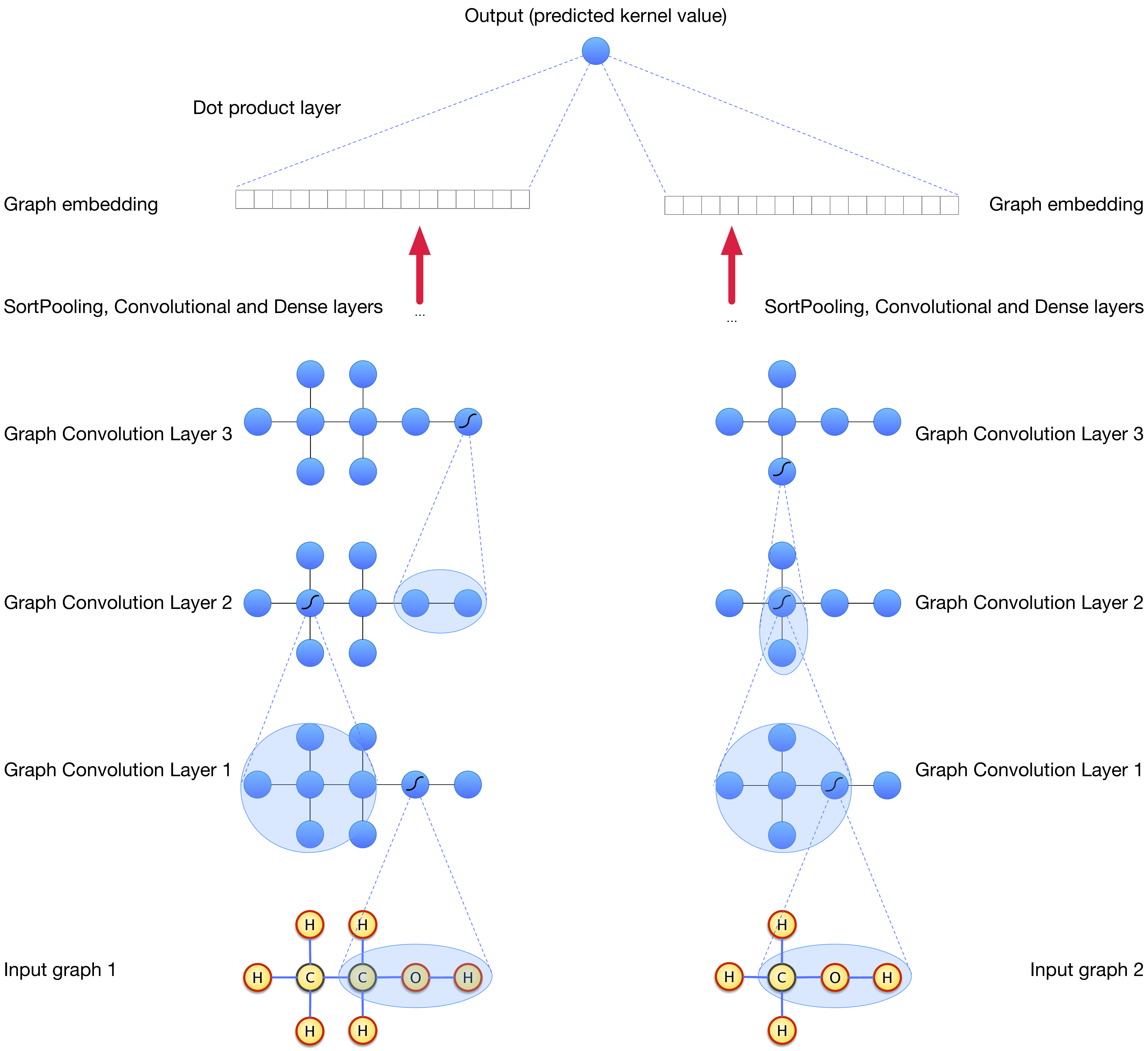}
    \caption{Our proposed siamese network for pre-training. Note that the parameters of the left and right branches are shared. We train the network providing as output the kernel value (computed with any graph kernel) between the two input molecules.}
    \label{fig:siamesecnn}
\end{figure}

We adopted as base network (the network that computes the $f$ function) the DGCNN model discussed in Section~\ref{sec:dgcnn}, removing the output layer.
Note that any model discussed in Section~\ref{sec:graphconvolution} can be adopted.
We learn the weights of the network via back-propagation.
Figure~\ref{fig:siamesecnn} summarizes our pre-training proposal.

After the pre-training phase, we remove the output layer of the siamese network, and we attach to one of the branches an output layer suited for the learning task in $D$. We can finally train the resulting network via back-propagation.

\section{Experimental results}
 We report experimental results on three datasets containing biological node-labeled graphs, namely MUTAG~\citep{Debnath1991}, PTC~\citep{Toivonen2003}, NCI1~\citep{springerlink:10.1007/s10115-007-0103-5}, Each graph represents a chemical compound, where nodes are labeled with the atom type, and edges represent bonds between them.
MUTAG is a dataset of aromatic and hetero-aromatic nitro compounds, where the task is to predict their mutagenic effect on a bacterium.
In PTC, the task is to predict chemical compounds carcinogenicity for male and female rats. NCI1 contains anti-cancer screens for cell lung cancer.

We compare the performance of our pre-trained method (Pre-trained DGCNN) to the same graph neural network without pre-training (DGCNN). We report, for sake of comparison, the performance of four state-of-the-art graph kernels: the graphlet kernel (GK) \citep{shervashidze2009efficient}, the random walk kernel (RW) \citep{vishwanathan2010graph}, the propagation kernel (PK) \citep{neumann2012efficient}, and the Weisfeiler-Lehman subtree kernel (WL) \citep{shervashidze2011weisfeiler}.
We adopted the WL kernel for our pre-training. However, in an extended version of this paper, we plan to study if the behavior of the proposed pre-training approach remains consistent also adopting other kernels. 

As for the unlabeled data used for pre-trianing, for sake of simplicity we decided to use the dataset itself in the pre-training phase. In the future, we plan to adopt other data (e.g. molecular datasets available from the NCI) for the pre-training phase.
%
\subsection{Experimental setup}
To evaluate the different methods, a nested 10-fold cross-validation is employed, i.e, one fold for testing, 9 folds for training of which one is used as validation set for model selection. For each dataset, we repeated each experiment 10 times and report the average accuracy over the 100 resulting folds. To select the best model, the hyper-parameters' values of different kernels are set as follows: the height of WL and PK in $\left\lbrace 0, 1, 2, 3, 4, 5 \right\rbrace$, the bin width of PK to 0.001, the size of the graphlets in GK to 3 and the  decay of RW to the largest power of 10 that is smaller than the reciprocal of the squared maximum node degree. Note that some of our results are reported from~\cite{Zhang2018}.
We employ for our siamese network the architecture presented in \cite{Zhang2018} (DGCNN). 
The activation function for the graph convolutions is the \emph{hyperbolic tangent}, while the 1D convolutions and the dense layer use \emph{rectified linear units}. 
%
The outputs of the two last dense layers (one for each branch of the network) are the input to a layer that computes the dot product between the two vectors.
We adopted as target for pre-training the WL kernel with a fixed number of iterations $h=2$.
We fixed the number of pre-training epochs to $20$ for MUTAG and PTC datastes, and to $2$ for NCI1 (for time limitations).
We then fine-tuned the pre-trained DGCNN as usual on the training dataset.

We used \emph{PyTorch}~\citep{paszke2017automatic} for our implementation.
We trained the Siamese network in an end-to-end fashion using Stochastic Gradient Descent with adaptive learning rate (\textit{adam}), using \emph{mean square error} as loss function for the training of the siamese network (pre-training), and \textit{negative log likelihood} as the loss function for the supervised phase.

\subsection{Experimental Results}
In Table~\ref{tab:reuslts-kernels} we report the results of our experiments.
On the MUTAG dataset, DGCNN is already the best-performing method among the considered ones. 
We recall that for our pre-training approach, we are using the WL kernel, that performs slightly worse than DGCNN on this dataset. It is interesting to see that in this case, our pre-training apporach can still improve over the random initialization (DGCNN). This result is an indicator that, even when the considered kernel is not the best performing method, it can still be used for providing insights to the training procedure.

As for the PTC dataset, the scenario is similar, having DGCNN that performs slightly better than WL, and with the pre-trained DGCNN performing better than its randomly initialized counterpart. In this case, PK is the best performing kernel. Even though pre-trained DGCNN performs better than PK, it would be interesting to see if using it for pre-training would improve the resulting performances even further. We leave this study for the future.

Finally, on NCI1 dataset, the WL kernel outperforms DGCNN accuracy of almost 10\%. Our pre-training approach improves over DGCNN as expected, but it is still far from the kernel's performance. We argue that this may be due to our choices for the pre-training hyper parameters, such as the low number of pre-training epochs that we adopted for this dataset (just 2, for time limitations), or the WL kernel iterations $h=2$, that is sub-optimal. In an extended version of this paper, we will extend our experimental results to understand if it is possible to close this gap.

\begin{table*}[t]
\def\arraystretch{1.6}
\begin{center}
\caption{Comparison with graph kernels.}
\label{tab:reuslts-kernels}
\begin{tabular}{lccc}
\hline
\textbf{Method / Dataset} & \textbf{MUTAG} & \textbf{PTC} & \textbf{NCI1}  \\
\hline
GK    & 81.39$\pm$1.74 & 55.65$\pm$0.46 & 62.49$\pm$0.27  \\ 
RW    & 79.17$\pm$2.07 & 55.91$\pm$0.32 & $>$3 days \\
PK    & 76.00$\pm$2.69 & 59.50$\pm$2.44 & 82.54$\pm$0.47  \\
WL    & 84.11$\pm$1.91 & 57.97$\pm$2.49 & \textbf{84.46$\pm$0.45}  \\
\hline
DGCNN & 85.83$\pm$1.66 & 58.59$\pm$2.47 & 74.44$\pm$0.47  \\
\hline
Pre-trained DGCNN  & \textbf{88.10$\pm$1.05} & \textbf{61.03$\pm$2.86} & 77.13$\pm$0.45 \\
\hline

\end{tabular}
\end{center}
\end{table*}

\section{Conclusions and Future Works}
In this paper, we proposed a pre-training technique for graph neural networks that exploits the graph kernels available in literature to define a target for a siamese network. The weights learned in this phase are used as initialization for the training phase on the actual target.

Since this is a preliminary work, we see several possibilities of extension.
One interesting route is to predict more kernels instead of just one. In this way, we are forcing the network to retain the information considered by different kernels. 
Another point is to verify if, with pre-training, we would be able to train deeper graph neural networks, thus further improving the predictive performance of such models.
%
%
Finally, we would like to investigate to what extent this approach is beneficial for other types of structured data.

\bibliographystyle{named}
\bibliography{references}
\end{document}